%% file: main.tex
\documentclass[runningheads,a4paper]{llncs}

\usepackage{amssymb}
\usepackage{graphicx}
\usepackage{url}
\usepackage[vlined]{algorithm2e}
\usepackage{amsmath}
\usepackage{csquotes}
\usepackage{coverpage}

\newcommand{\keywords}[1]{\par\addvspace\baselineskip
\noindent\keywordname\enspace\ignorespaces#1}

\begin{document}

\mainmatter  

\title{An integrated approach to autonomous environment modeling}
\titlerunning{An integrated approach to autonomous environment modeling}

\author{Miroslav Kulich \and Viktor Koz\'ak \and Libor P\v{r}eu\v{c}il%
}


\institute{
Czech Institute of Informatics, Robotics, and Cybernetics,\\
Czech Technical University in Prague, Prague, Czech Republic\\
kulich@cvut.cz, viktor.kozak@cvut.cz, preucil@cvut.cz\\
\url{http://imr.ciir.cvut.cz}}

%
%


\mauthor{Miroslav Kulich, Viktor Koz\'ak and Libor P\v{r}eu\v{c}il}
\published{{\it Modelling and Simulation for Autonomous Systems (MESAS 2017)}, Cham: Springer International Publishing AG, 2018. p. 3-17. ISBN 978-3-319-76071-1.}
\DOI{10.1007/978-3-319-76072-8\_1}
\original{https://link.springer.com/chapter/10.1007/978-3-319-76072-8\_1}
\coverpage

\maketitle

\begin{abstract}
In this paper, we present an integrated solution to memory-efficient environment modeling by an autonomous mobile robot equipped with a laser range-finder. 

Majority of nowadays approaches to autonomous environment modelling, called exploration, employs occupancy grids as environment representation where the working space is divided into small cells each storing information about the corresponding piece of the environment in the form of a probabilistic estimate of its state. In contrast, the presented approach uses a polygonal representation of the explored environment which consumes much less memory, enables fast planning and decision-making algorithms and it is thus reliable for large-scale environments. 

Simultaneous localization and mapping (SLAM) has been integrated into the presented framework to correct odometry errors and to provide accurate position estimates. This involves also refinement of the already generated environment model in case of loop closure, i.e. when the robot detects that it revisited an already explored place.

The framework has been implemented in Robot Operating System (ROS) and tested with a real robot in various environments. 
The experiments show that the polygonal representation with SLAM integrated can be used in the real world as it is fast, memory efficient and accurate. Moreover, the refinement can be executed in real-time during the exploration process.

\end{abstract}

\input{src/intro}
\input{src/approach}
\input{src/polygonal}
\input{src/loc}

\input{src/experiments}
\input{src/conclusion}

\section*{Acknowledgement}
This work has been supported by the European Union's Horizon 2020 research and innovation programme under grant agreement No 688117, by the Technology Agency of the Czech Republic under the project no.~TE01020197 \enquote{Centre for Applied Cybernetics}, the project Rob4Ind4.0 CZ.02.1.01/0.0/0.0/15\_003/0000470 and the European Regional Development Fund.
The work of T. Juchelka and V. Lhotsk\'y on the previous versions of the exploration framework is also highly appreciated.

\bibliographystyle{splncs03}
\bibliography{main}

\end{document}

%% file: src/intro.tex
\section{Introduction}
\noindent 
Knowledge of a model of the working area of an autonomous system is a necessary condition to make qualified decisions about current and future actions. 
The more accurate the model is, the better decisions can be made and thus increase performance of the system. Unfortunately, a priori knowledge of the environment is not available in many applications and has to be acquired from scratch. 
Exploration -- the process of autonomous environment modelling is generally defined as an iterative procedure consisting of several steps. Actual sensory information is read first and the model of the environment is updated after some data processing and noise filtering. A new goal for the robot is then determined, the shortest path to it is found along which the robot is navigated. These steps are repeated until no unexplored area remains. 

In this paper, several assumption are made in order to simplify the problem and to focus on representation of the actual knowledge about the environment in which the robot operates.
We particularly assume that the robot is equipped with a laser range finder and it operates in 2D.

The most popular approach to exploration is frontier-based exploration introduced by Yamauchi~\cite{Yamauchi98} and further extended by many researchers~\cite{wurm08,burgardCoordinated,Amigoni08,holz10,Karpov2016}.
The approach is based on occupancy grids where the working space is divided into small cells and each cell stores information about the corresponding piece of the environment in the form of a probabilistic estimate of its state.

Several authors do not build an exact metric map.
Instead, they incrementally create topological information about the space in the form of a graph.
Frontier-based modification of {\em Sensor-based Random Tree}, a probabilistic strategy, which represents a roadmap of the explored area with an associated safe region is presented in~\cite{freda05}. 
The approach has been generalized in~\cite{franchi09}, where {\em Sensor-based Random Graph} is constructed.
Feature-based map is used in~\cite{newman03}.
Moreover, a free space is represented by a set of so-called markers, which are connected based on visibility constrain.

Combination of metric (in the form of occupancy grid) and topological maps is presented in~\cite{poncela02,Jia2012}. 
The metric map is built first, while a hierarchical structure is created over it leading to topological map construction.
The opposite approach ({\em Spatial Semantic Hierarchy}) defines distinctive places and paths in order to build a topological description, while geometric knowledge is  assimilated onto the elements of this description~\cite{kuipers91}.  
Another combination of metric and topological maps is introduced in~\cite{zhang2015}, where a generalized Voronoi Graph built from preprocessed raw laser measurement is employed as a topological map for path planning, goal selection, and SLAM.  
Finally, a flux-based skeletonization algorithm on the latest occupancy grid is employed for on-line construction of a topological map for exploration in~\cite{Rezanejad2015}.  

Shen et al.~\cite{Shen2012} propose a stochastic differential equation-based algorithm. 
They use a system of particles with Newtonian dynamics to determine regions for further exploration in 3D for unmanned aerial vehicle.

Also, exploration based on a polygonal representation is not new, although it is used for a single robot only.
Gonz{\'a}lez-Ba{\~n}os and Latombe~\cite{Gonzalez-BanosL02} introduce a concept of a {\em Safe Region}, the largest region guaranteed to be obstacle-free given the history of sensor readings.
A map is iteratively built by executing union operations over successive safe regions.
{D}akulovi{\'c} et al.~\cite{djakulovic2011exploration} extend Ekman's approach~\cite{ekman97}. 
For each scan, a polygon is created using line-fitting on scan points, then Vatti's algorithm~\cite{vatti92} is used to compose particular polygons.
However, quantitative evaluation and performance comparison are missing in these papers, they present only few pictures with obtained polygonal maps.

The rest of the paper is organized as follows.
The problem definition is presented in Section~\ref{sec:problem}, while the approach itself is introduced in Section~\ref{sec:polygonal}. The integration of simultaneous localization and mapping for determination of robot position is  described in Section~\ref{sec:slam}.
Evaluation of the results and discussions are presented in Section~\ref{sec:experiments}.
Finally, Section~\ref{sec:conclusion} is dedicated to concluding remarks.

%% file: src/approach.tex
\section{Problem definition\label{sec:problem}}
Exploration is the process in which a robot autonomously operates in an unknown environment with the aim to built a map of it.
The map is built incrementally as actual sensor measurements are gathered and it serves as a model of the environment for further exploration steps.

The exploration algorithm consists of several steps that are repeated until some unexplored area remain.
The process starts with reading actual sensor information by the robots.
After some data processing, the existing map is updated with this information.
New goal candidates are then determined and a new goal to be visited by the robot is determined using a defined cost function.

Having determined the goal, the shortest path from robot position to the goal is found.
Finally, the robot is navigated along the path.
The whole exploration process is summarized in Algorithm~\ref{sc:exploration}.

\begin{algorithm}[ht]
\While{unexplored areas exist}{
  read current sensor information\;
  update map with the obtained data\;
  determine new goal candidates\;
  determine the new goal\;
  plan a path to the goal\;
  move the robot towards the goal\;
}
\caption{The exploration algorithm}
\label{sc:exploration}
\end{algorithm}

In this paper, we follow Yamauchi's frontier based approach~\cite{Yamauchi98}, which assumes that the next best view (goal) lies on the border between free and unexplored areas (this border is called {\em frontier}).

%% file: src/polygonal.tex
\section{Polygonal domain}
\label{sec:polygonal}
In the presented approach, the information about the environment is approximated by a polygon with holes (i.e., the outer polygon representing a border of the working area and containing obstacles -- holes). 
This polygon $\cal P$ is, similarly to~\cite{djakulovic2011exploration}, incrementally created as a union of polygons ${\cal P}_i$  representing sensor measurements (scans) taken during the mission: ${\cal P}=\cup_{i=0}^{t} {\cal P}_i$, where $t$ is the actual time.

The particular polygon ${\cal P}_i$ is created from a range data ${\cal R}_i$ that are typically represented as a vector of points. 
This is a standard task and many approaches for polygon building from sensory data have been developed.
The combination of three algorithms is used:
\begin{itemize}
\item {\em Successive Edge Following}~\cite{siadat97}  splits scan data into clusters representing particular objects,
\item {\em Ramer--Douglas--Peucker} algorithm~\cite{hershberger92} smooths objects' boundaries into piecewise linear curves (polylines),  
\item {\em Least Squares Fit} finds parameters of lines to best fit the scan data. 
\end{itemize}
Finally, the position of the sensor is added between the first and the last point and successive polylines are connected.
The resulting polygon represents a free space as detected by the measurement. 
We distinguish between two types of edges: those representing an obstacle and those that were added in the last step, which represent frontiers.

\begin{figure}[ht]
\centering
\includegraphics[width=0.9\columnwidth]{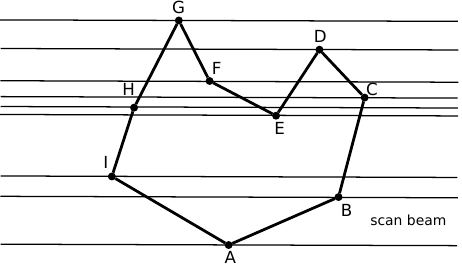}
\caption{Vatti's representation of a polygon. The vertices \textit{A,B,C,D} form the right bound and the vertices \textit{A,I,H,G} form the left bound. The vertex \textit{A} is the local minimum and the vertex \textit{D} is the local maximum. }
\label{fig:vatti}
\end{figure}

The union of polygons obtained from different measurement poses can be computed by Vatti clipping algorithm~\cite{vatti92}\footnote{In our implementation, we use the Clipper library~\cite{clip}, which is an open-source polygon clipping library based on Vatti clipping algorithm.  The library performs the boolean clipping operations - intersection, union, difference, and XOR. Moreover, it performs polygon offsetting.}.
The algorithm can handle large sets of polygons, polygons with holes, and self-intersecting polygons.
On the other hand, adding information about edge type is not straightforward since this information has to be preserved by clipping operations.

The Vatti algorithm processes both involved polygons by a sweep line starting at the lowermost vertex and going upwards passing through all vertices of the polygons.
Bounds --- sequences of consecutive edges starting at a local minimum and ending at a local maximum --- are formed during this process. 
Left and right bounds are distinguished with respect to their positions to the polygon's interior.
A polygon described with this notation is shown in Fig.~\ref{fig:vatti}

\subsection{Modifications of clipping}
\begin{figure}[ht]
\centering
\includegraphics[width=0.6\columnwidth]{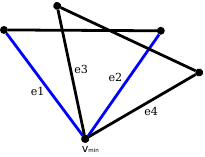}
\caption{Problem in a local minimum.}
\label{fig:lm}
\vspace{-1em}
\end{figure}

The straightforward approach how to add and preserve information about the edge type lies in modifications of specific parts of the Vatti algorithm.
In this case, information about the edge type is stored in vertices adjacent to it.
Unfortunately, this approach fails.
One of the main reasons is that a relation between output vertices and the input edges can not be determined easily.
For example, if two different bounds share their local minimum, a new vertex is added into the output polygon.
There is no guarantee which bound the algorithm takes first.
When the second bound is processed, its vertex is skipped, because it was already added.
The example situation is illustrated in Fig.~\ref{fig:lm}, where $v_{min}$ is the local minimum which is added to the output either in processing edges $e_1$ and $e_2$ or $e_3$ and $e_4$.
Because the first pair of edges is a frontier, while the second one is not, the parameters of the added vertex may be different.

\begin{figure}[t]
\centering
\includegraphics[width=0.5\columnwidth]{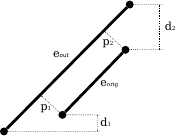}
\caption{Comparison of edges.}
\label{fig:crit}
\end{figure}

Our approach post-processes edges of the output polygon, compares them with edges of the input polygons and assigns them the correct type.
The comparison of each output edge is made by computing a penalty function for all edges in the input polygons.
The penalty value can be expressed as the sum of all the distances depicted in Fig.~\ref{fig:crit}.
\begin{equation}
P = p_1 + p_2 + |d_1| + |d_2|
\end{equation}
The distances $p_1$,$p_2$ are the perpendicular distances from vertices of $e_{orig}$ to $e_{out}$ and $d_1$,$d_2$ are differences in $y$-axis.
Notice that the distance $d_1$ is considered only if the bottom vertex of $e_{out}$ has lower $y$-coordinate than the bottom vertex of $e_{orig}$ and the distance $d_2$ is considered only if the top vertex of $e_{out}$ has higher $y$-coordinate than the top vertex of $e_{orig}$.
The best input edge, i.e., the edge with the lowest penalty, is considered as correct and the information from the found input edge is copied into the output edge.

\begin{figure}[ht]
\centering
\includegraphics[width=\columnwidth]{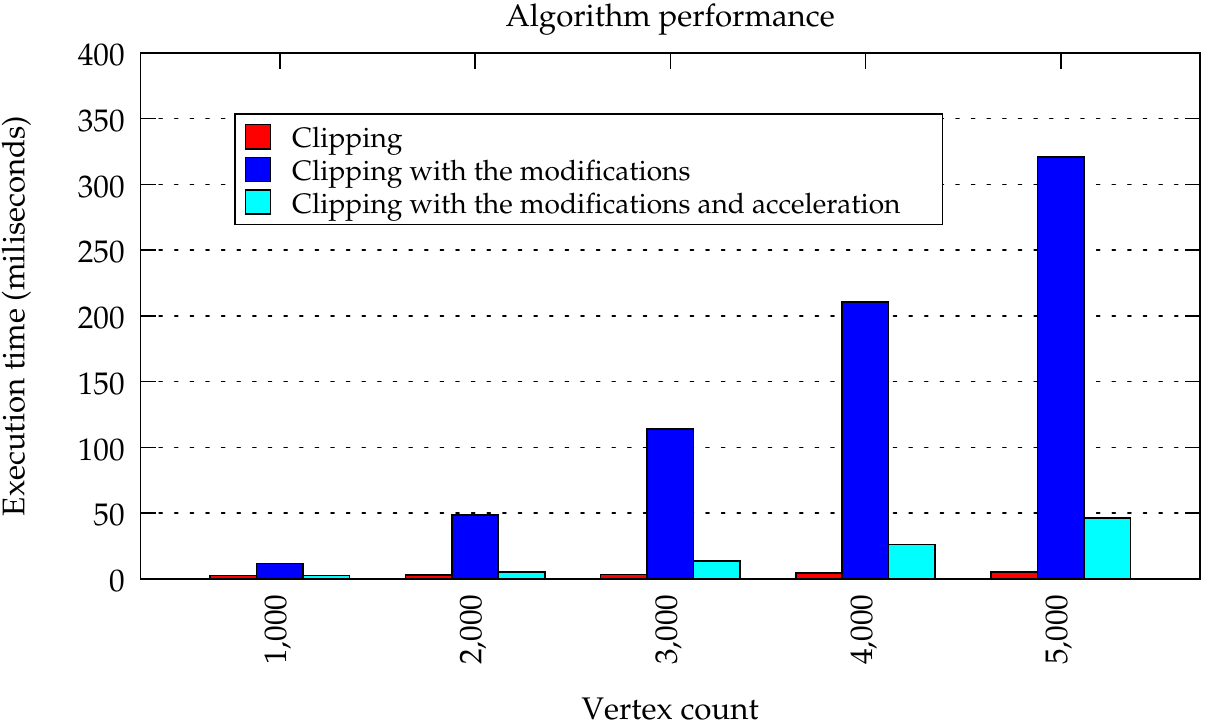}
\caption{Performance of the modified algorithm.}
\label{fig:speeds}
\end{figure}

This process is much more computationally complex than the clipping algorithm itself.
From the knowledge about the clipping algorithm it is possible to do some simplifications that speed up the matching.
The clipping algorithm creates the bounds with the edges in a bottom-up fashion starting at the local minimum.
The most important fact is that the edges are ordered by $y$-coordinate.
The bounds are also ordered by $y$-coordinate of their local minimum.
These internal structures can be used instead of the original polygons.
The edges of the input polygons are processed based on their order in bounds while the following rules are applied:
\begin{itemize}
\item {The bounds with $y$-coordinate of its local minimum higher than $y$-coordinate of the top vertex of the output edge can be completely skipped.} 
\item {If $y$-coordinate of the bottom vertex of the edge from the bound is higher than the top vertex of the output edge then the rest of one bound can be skipped.}
\item {If the penalty value is zero then skip further comparison of the output edge.}
\end{itemize}
These criteria improve the speed of the algorithm significantly.
Fig.~\ref{fig:speeds} shows how the simplifications affect the algorithm performance.
Although the modifications even with simplifications slow-down the clipping, the approach can be applied on real problems.
Maps containing $1000$ vertices are processed in few milliseconds.

\subsection{Polygon offsetting}
The map created by the clipping algorithm is useful for path planning for a point-robot only. 
If a robot is approximated by a disk, the map (i.e., each obstacle) has to be enlarged by constructing the Minkowski sum of the map with this disk. 
This can be done by polygon offsetting operation.  
Management of the edge type is similar to the clipping process.
Notice that enlarging can lead to intersecting or self-intersecting polygons and therefore the same post-processing as for polygon clipping is used.

\subsection{Map representation}
\label{sec:repre}
It is possible to maintain the knowledge about the environment as the all-in-one map, but this is not robust.
Both data from a laser range finder and odometry can be noisy, which causes that some frontiers may be generated nearby or inside obstacles.
The more robust approach is to represent the maps of a free-space and obstacles separately.
Whenever a new scan is added into the map it is added into the free-space map as it is.
The scan is next checked whether it contains obstacles.
If so, the obstacles are offset (proportionally to the map size and noise) and added into the obstacle map.

Before the map is used for planning, both maps are temporarily combined together into a single map.
The modified clipping is performed here and the resulting map contains the information about frontiers.
If the resulting edge comes from the free-space map it is considered as a frontier and it is marked as an obstacle otherwise.

The created map contains a large number of vertices. 
Ramer--Douglas--Peucker algorithm is thus used after each clipping or offsetting in order to reduce this number. 
 
The map is used for goal candidates selection, planning, and evaluation of the candidates in each exploration step.
The selection process determines the candidates so that they lie on frontiers, the distance of a candidate to its neighbors is twice a sensor range, and all frontiers are inside the union of circles with centers in the candidates and radius equal to the sensor range.
This guaranties that all frontiers will be explored (i.e., it will be detected whether a frontier lies in a free space or in any obstacle) after visiting all candidates.

Planning consists of two steps: a visibility graph is computed first, followed by several runs of Dijkstra algorithm, which computes shortest distances among each robot and a goal candidate.
The cost function for evaluation of goal candidates is then simply this distance.


%% file: src/loc.tex
\section{Simultaneous localization and mapping}
\label{sec:slam}
The presented mapping approach assumes that a precise position of the robot is known.
Unfortunately, this is not a case in many real-world applications, so robot position has to be determined during a mission based on data from sensors. 
The process of simultaneous localization and mapping  (SLAM) is a well know problem for which many algorithms have been developed in the last decades.
One of the most popular approaches is a particle filter~\cite{grisetti2007fast,montemerlo2007fastslam} which represents probability density function of robot position by a set of particles, each storing a single estimate of robot trajectory together with a map built along this trajectory.
More precisely, the particle filter computes an accurate distribution of particles taking into account movement of the robot and the most recent observation. 
The key idea is to estimate a posterior $p(x_{1:t} | z_{1:t},u_{0:t})$ about probable trajectories $x_{1:t}$ of the robot given its observations $z_{1:t}$ and its odometry measurements $u_{0:t}$. This posterior is further used to compute a posterior over maps and trajectories:

\begin{equation} \label{posterior}
p(x_{1:t}, m | z_{1:t},u_{0:t}) = p(m | x_{1:t},z_{1:t})p(x_{1:t} | z_{1:t},u_{0:t})
\end{equation}

The process of the SLAM algorithm iteratively creates a map of the environment and updates positions of the particles. Every iteration consists of five main steps~\cite{grisetti2007fast}:
\begin{enumerate} 
\item \textit{Sampling}: New generation of particles $x_t^{(i)}$ is obtained from the current generation $x_{t-1}^{(i)}$ by sampling from a proposal distribution $\pi(x_t | z_{1:t}, u_{0:t})$.

\item \textit{Importance weighting}: Each particle is assigned an individual importance weight $w^{(i)}$ according to:\begin{equation}
w^{(i)} = \frac{p(x_t^{(i)} | z_{1:t})}{\pi(x_t^{(i)} | z_{1:t})}
\end{equation}
These weights account for the fact that the proposal distribution $\pi$ is in general not equal to the target distribution of successor states.

\item \textit{Resampling}: Particles with low weights are replaced by samples with higher weights. Due to a limited number of particles it is important to approximate a continuous distribution. Resampling also allows to apply a particle filter in situations in which the true distribution differs from the proposal.
\item \textit{Map Estimation}: For each particle position $x_{t}^{(i)}$ is computed a corresponding map estimate $m_{t}^{(i)}$, based on the previous trajectory and the history of observations according to $p(m_{1:t}^{(i)} | x_{1:t}^{(i)}, z_{1:t})$.
\item \textit{Output selection}: 
\label{sec:best_particle}
The particle with the highest importance weight is determined as the \textit{best} particle and it's position is published as the current position of the robot, together with its corresponding map estimate.
\end{enumerate}

\subsection{Integration into the exploration framework for a polygonal domain}
A straightforward way how to integrate SLAM into realization of the exploration process is to use an existing implementation of SLAM. 
While many SLAM libraries can be found, none of them works in a polygonal domain.
We therefore employed the GMapping library~\cite{gmapping} which uses grids maps as a map representation.

A naive approach to SLAM integration with a polygonal map building process is to take robot position from GMapping as a position of current sensor readings and update a polygonal map making use of this information.   
Nevertheless, updating robot position only can lead to big inconsistencies in the map especially when the robot revisits an already explored place (loop closure). 
In case of loop closure, a large error in position estimation can be detected and corrected by the SLAM algorithm. 
This is done by switching a best particle in step~\ref{sec:best_particle} of the SLAM algorithm, which results in a new estimate of a complete robot trajectory $x_{1:t}$ and thus a map.

It would be unreasonable to perform the refinement of a polygonal map every time a new best particle is selected. 
The refinement would not only require information on all previous particles and their positions, but more importantly it would be triggered whenever a switch of the best particle occurs, resulting in the refinement on frequent occasions even without closing the loop. 
Since each change could lead to a refinement of the map a few dozens or even hundreds iterations back, a different approach is used.

Individual polygonal maps are continuously created for each particle instead and the map belonging to the best particle is published as the resulting map and used for the exploration. 
Since polygonal maps require a small amount of memory, keeping several dozens of them at once has little effect on the memory requirements. 
During the experiments the control unit also proved to have no issue with the increase in computational complexity caused by the processing of multiple polygonal maps every iteration.

%% file: src/experiments.tex
\section{Experiments}
\label{sec:experiments}

The presented exploration framework has been implemented int Robot Operating System (ROS)~\cite{Ros09} and several experiments were made to evaluate functionality of the system.
The experiments were performed making use a TurtleBot mobile robot equipped with the SICK LMS 111-10100 laser rangefinder. 
All the algorithms and computations were run on an INTEL NUC5i5RYK control unit placed on the robot with a dual core 1.6 GHz processor, 16 GB RAM and running the Ubuntu 14.04 operation system. 

\begin{figure}[htbp]
\centerline{ 
\includegraphics[width=0.49\textwidth]{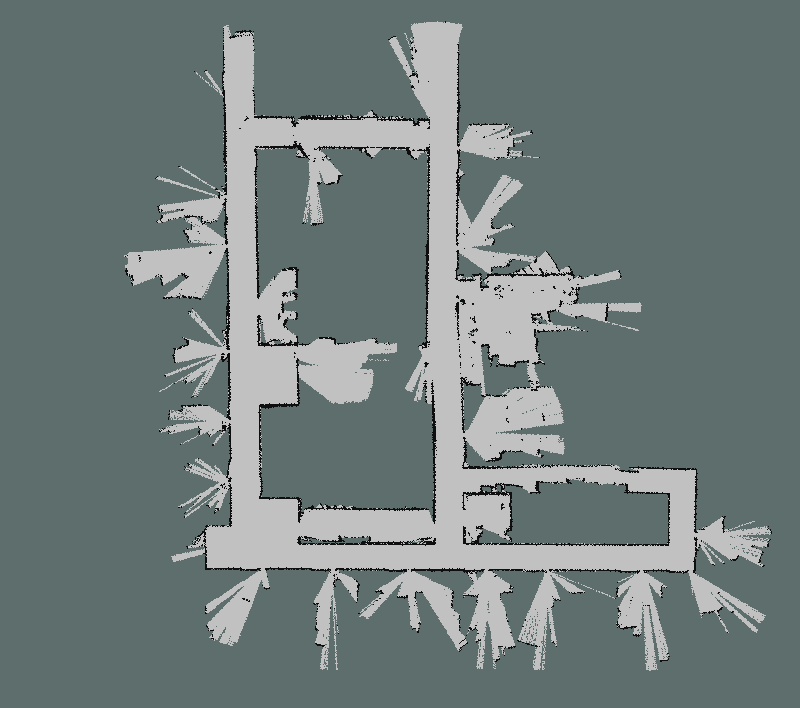}
    \hskip 0.5 cm
\includegraphics[width=0.49\textwidth]{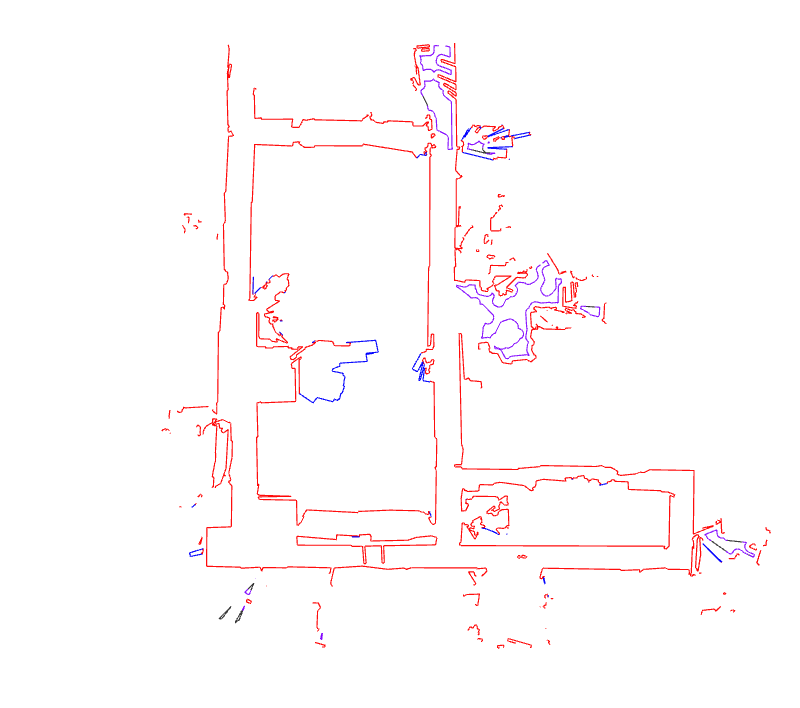}
    } 
  \nobreak\medskip
  \caption{A map of the office-like environment created by the exploration algorithm. The GMapping occupancy grid map is on the left and the created polygonal map is on the right.}   
\label{fig:blox_slam}
\end{figure}

The first part of the experiments, where GMapping is used to provide pose estimates only, was executed in indoor offices. 
The robot starts its movement in one of the office rooms and explores narrow corridors and adjacent resting areas. 
Fig. \ref{fig:blox_slam} shows a grid map created by the GMapping SLAM algorithm and the final polygonal map created by the EAPD package, a clear correspondence with the map from GMapping can be seen.

The robot had to deal with a lot of passers-by in the building. 
The exploration algorithm had no problem with such environment and both the map and exploration algorithm were able to recover within a short time after encountering a temporary obstacle in the environment.

\begin{figure}[htbp]
\centerline{ 
\includegraphics[width=0.49\textwidth]{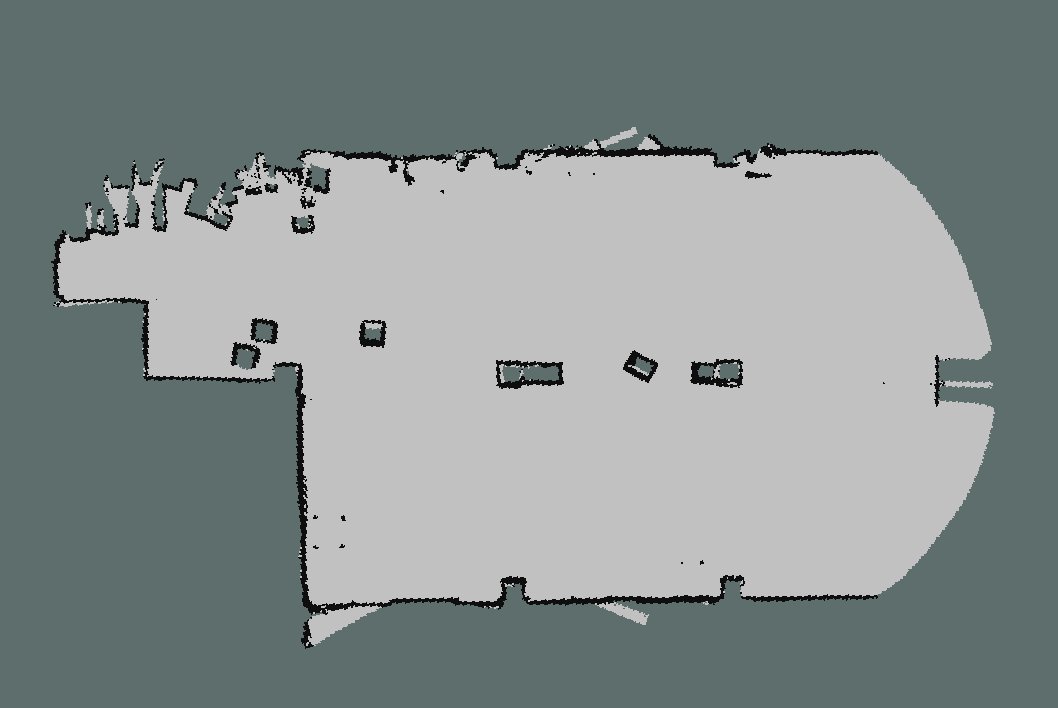}
    \hskip 0.5 cm
\includegraphics[width=0.49\textwidth]{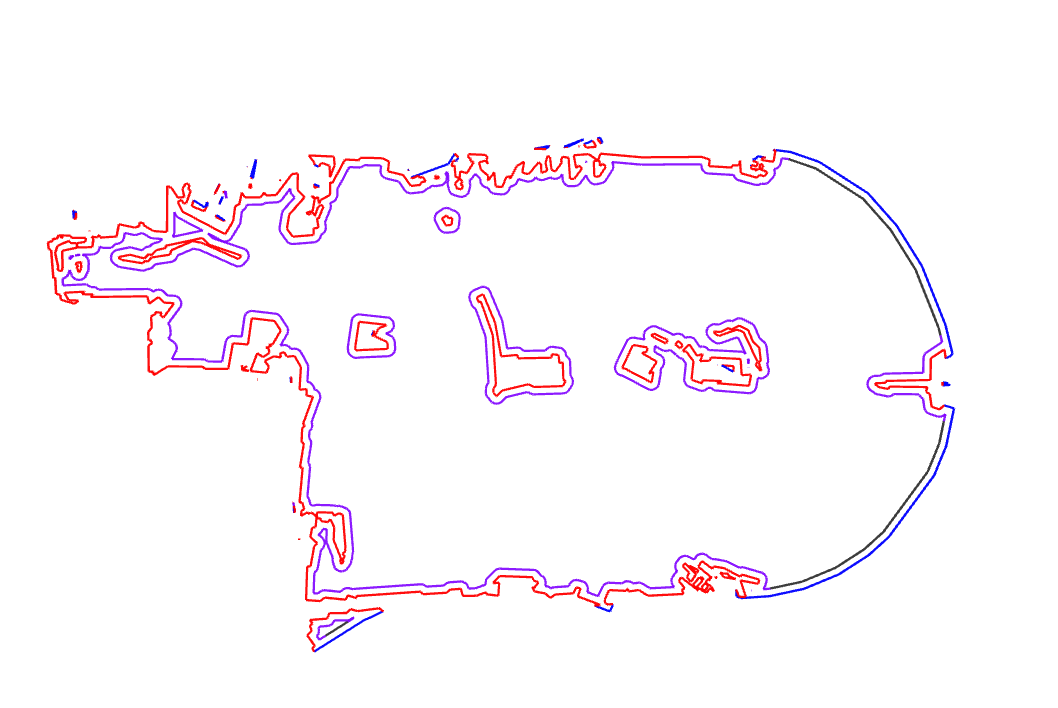}
    }
  \nobreak\medskip
  \caption{A map of the laboratory created by the exploration algorithm. The GMapping occupancy grid map is on the left and the created polygonal map is on the right.}   
\label{fig:ciirk_slam}
\end{figure}

Other experiments were made in a large open space laboratory, which  provides a different challenge than narrow corridors. 
The robot had to face the difficulty of navigation through an open environment. 
Situations, where most obstacles are out of the range of the laser sensor, are challenging for the laser-based SLAM algorithm and the robot has an increased need for a precise odometry information. 
Fig. \ref{fig:ciirk_slam} presents the created maps.

The functionality of the system is apparent, as the exploration was fully autonomous and the robot was able to create a model of the environment in both buildings. 

Section \ref{sec:slam} presents the closer integration of GMapping with the polygonal exploration. The GMapping package was modified to publish specific information on all particles allowing the polygonal map building process to synchronize with the SLAM algorithm. 
This synchronization allows the adaptation to the changes in the best particle. 
Description of experiments testing this functionality follows.

\begin{figure}[htbp]
\centerline{ 
\includegraphics[width=0.49\textwidth]{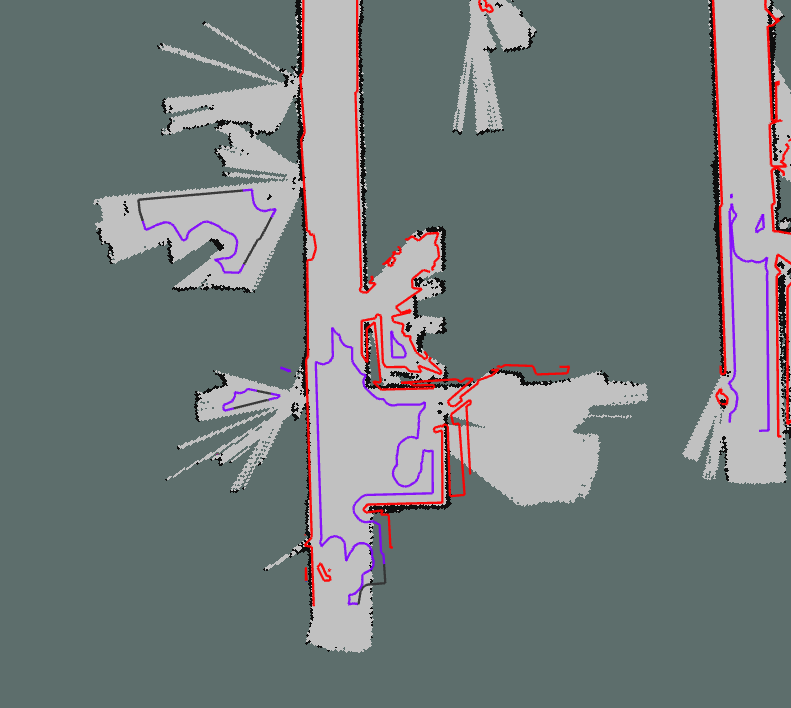}
    \hskip 0.5 cm
\includegraphics[width=0.49\textwidth]{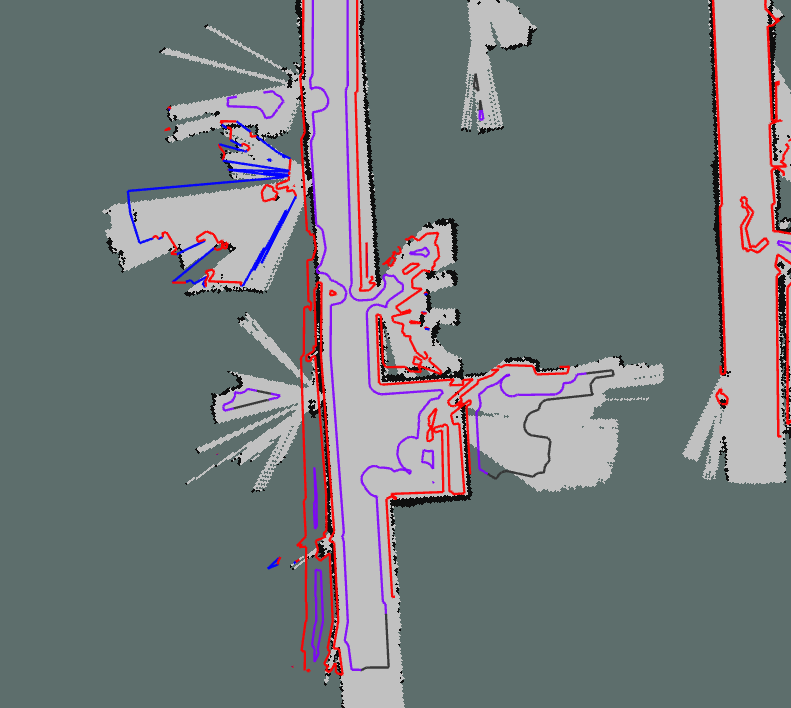}
    } 
  \nobreak\medskip
  \caption{An overlay of the polygonal map and the grid map, before(left) and after(right) a switch in the \textit{best} particle in the GMapping SLAM algorithm. The situation occurred during the experiments testing the slam functionality, connected with the creation of the map in Fig. \ref{fig:blox_slam}. }
\label{fig:blox_switch}
\end{figure}

Although the SLAM algorithm proved functional during the previous experiments, its connection with the exploration package was limited only to the estimates of the current position. 
This resulted in the inability of the exploration algorithm to properly react to the changes made by the GMapping SLAM during a switch between particles. 
Fig. \ref{fig:blox_switch} presents a situation shortly before and after the switch in the best particle. The image on the left presents the state prior to the particle switch, while the image on the right shows an inconsistency in maps produced by the algorithms, caused by a change in the best particle.

The GMapping algorithm processes all particles during the process, it has already assimilated a full path of the particle leading to the current position and only switched between the particles and their map representations. 
However the polygonal mapping has been following a different particle up to the point where the switch occurred and received only the resulting change in position at the time of the switch, therefore the map created prior to the switch occurrence follows a different path than the path of the current best particle. 

\begin{figure}[htbp]
\centering
\includegraphics[width=\textwidth]{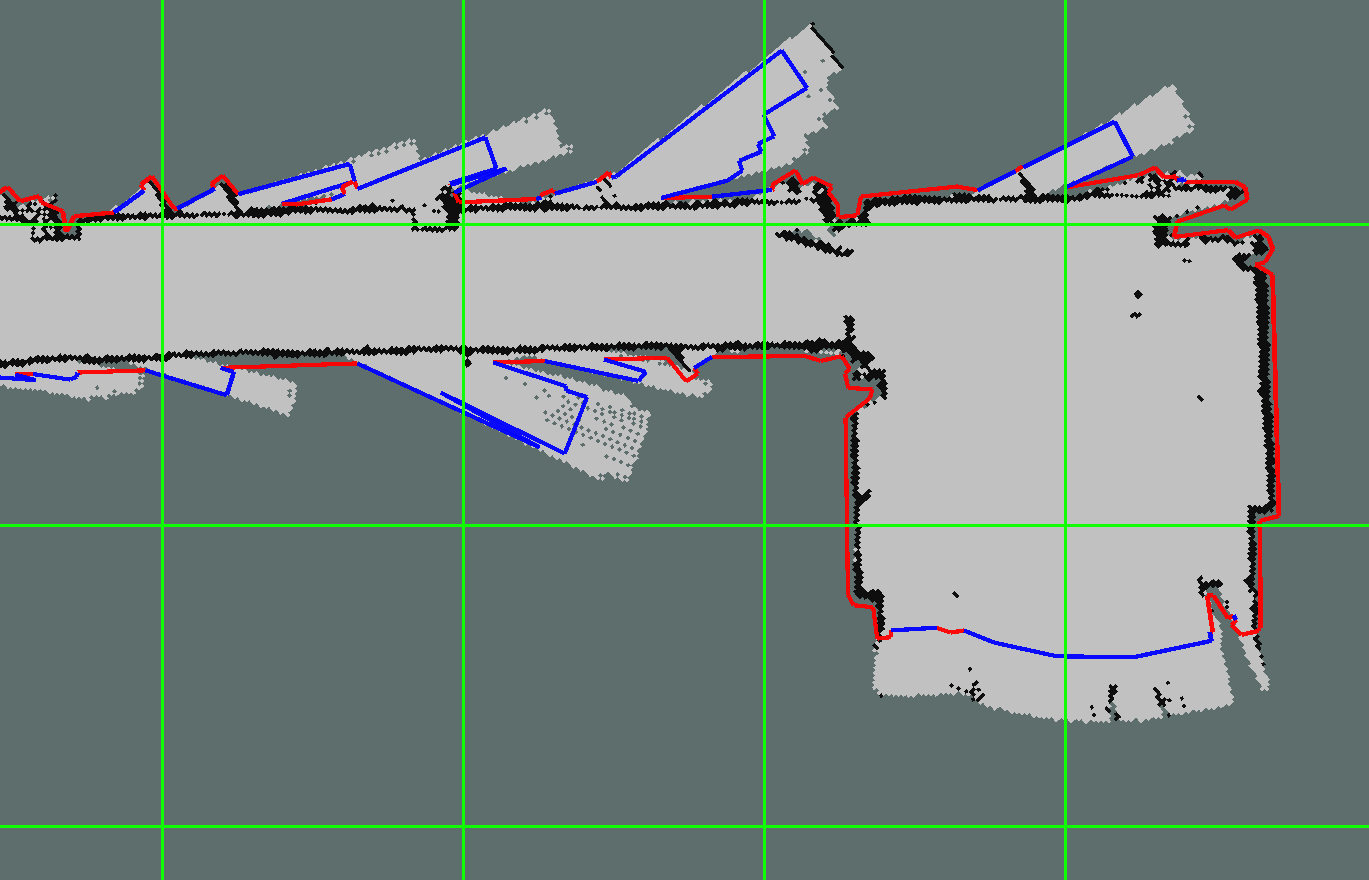}
\bigskip 
\includegraphics[width=\textwidth]{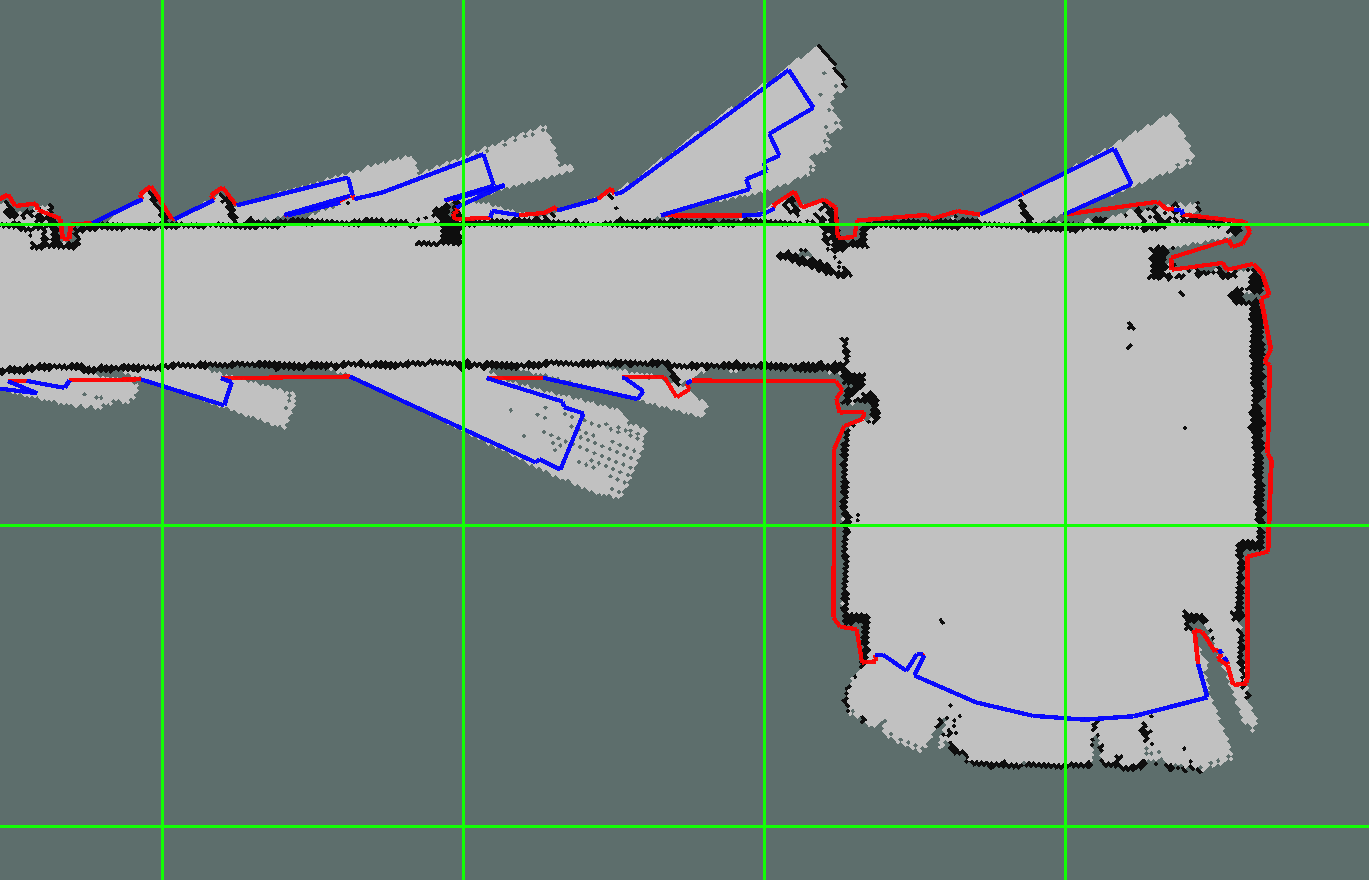}

  \caption{An overlay of the polygonal map and the grid map, before(top) and after(bottom) a switch in the \textit{best} particle in the GMapping SLAM algorithm. A green grid has been added to emphasize the shift in the structure of the maps.}
\label{fig:switch}
\end{figure}

The modifications made in section \ref{sec:slam} were initially proposed to integrate GMapping loop closure functionality, but since the functionality in GMapping projects itself in the selection of the best particle, the approach had to be changed and the exploration framework had to be extended to support switches between the particles. Experiments testing the new functionality were performed and are focused on the proper functionality during a particle switch.

Fig. \ref{fig:switch} presents the situation shortly before and after the switch in the \textit{best} particle. As can be seen in the figure, the structure of both maps shifted in reaction to the switch. The maps are fully aligned after the shift, which proves the correct functionality of the implementation.

%% file: src/conclusion.tex
\section{Conclusion}
\label{sec:conclusion}
The paper introduces a polygonal approach for autonomous environment modelling.
The mapping process is based on a standard library for polygon clipping, so it is robust and fast.
This enables to perform exploration in larger experiments, with higher number of robots, and make re-planning faster than possible in a grid-based approach. 
The most challenging task was to modify the clipping library in order to work with the frontiers.
After several unsuccessful attempts to change the clipping algorithm a compromise solution was found in the edge matching.  

SLAM functionality providing precise robot position was integrated into the exploration framework.
Specifically, the GMapping library was extended to store a polygonal map in each particle, which removes negative impact of jumps in the estimated position caused by a switch of the best particle in the SLAM algorithm, improving the precision of the created polygonal map. The method also supports the innate loop closure functionality of the GMapping library. The downside is that the new implementation has higher computational requirements than the original mapping process. 
Nevertheless, the performed experiments with a real robot show that the map refinement
can be executed in real-time during the exploration on moderately large segments of the map.

A big disadvantage of the current solution is that GMapping internally uses memory inefficient occupancy grids as a map representation which also leads to higher computational burden.
The future work will thus focus on full removal of occupancy grids from GMapping and their substitution with polygonal maps.